\documentclass[journal]{IEEEtran}

\usepackage[utf8]{inputenc}
\usepackage{amsmath, amssymb, amsfonts}
\usepackage{amsthm}
\usepackage{acronym}
\usepackage{graphicx}
\usepackage{subfigure}
\usepackage{graphicx}
\usepackage{subfigure}
\usepackage{cleveref}
\usepackage{algorithm,algorithmic}

\theoremstyle{plain}

\theoremstyle{definition}

\theoremstyle{remark}

\newcommand{\FIG}[1]{Fig.~\ref{#1}}
\newcommand{\SEC}[1]{Section~\ref{#1}}
\newcommand{\TAB}[1]{Table~\ref{#1}}

\DeclareMathOperator*{\argmin}{argmin}

\acrodef{CNN}{convolutional neural network}
\acrodef{ReLU}{rectified linear unit}
\acrodef{SGD}{stochastic gradient descent}
\acrodef{SVM}{support vector machine}

\begin{document}

\title{Multi-Task and Transfer Learning for Federated Learning Applications}

\author{Cihat~Ke\c{c}eci,
	Mohammad~Shaqfeh,~\IEEEmembership{Member,~IEEE,}
	Hayat~Mbayed,
	and~Erchin~Serpedin,~\IEEEmembership{Fellow,~IEEE}%
	\thanks{This publication was made possible by NPRP13S-0127-200182 from the Qatar National Research Fund (a member of Qatar Foundation). The statements made herein are solely the responsibility of the authors. (Corresponding author: Cihat Ke\c{c}eci.)}
	\thanks{Cihat Ke\c{c}eci and Erchin Serpedin are with the Department of Electrical and Computer Engineering, Texas A\&M University, College Station, TX 77843 USA (e-mail: \{kececi,eserpedin\}@tamu.edu).}%
	\thanks{Mohammad Shaqfeh and Hayat Mbayed are with the Department of Electrical and Computer Engineering, Texas A\&M University at Qatar, Doha, Qatar (e-mail: \{mohammad.shaqfeh,hayat.mbayed\}@qatar.tamu.edu).}%
}

\maketitle

\begin{abstract}
Federated learning enables many applications benefiting distributed and private datasets of a large number of potential data-holding clients. However, different clients usually have their own particular objectives in terms of the tasks to be learned from the data. So, supporting federated learning with meta-learning tools such as multi-task learning and transfer learning will help enlarge the set of potential applications of federated learning by letting clients of different but related tasks share task-agnostic models that can be then further updated and tailored by each individual client for its particular task. In a federated multi-task learning problem, the trained deep neural network model should be fine-tuned for the respective objective of each client while sharing some parameters for more generalizability. We propose to train a deep neural network model with more generalized layers closer to the input and more personalized layers to the output. We achieve that by introducing layer types such as pre-trained, common, task-specific, and personal layers. We provide simulation results to highlight particular scenarios in which meta-learning-based federated learning proves to be useful.
\end{abstract}


\section{Introduction}

Data privacy is of utmost importance with the increasing amount of digital data. The conventional deep learning methods require the collection and storage of the training data, which may contain sensitive information about the users.
At the same time,  modern data-oriented applications increasingly adopt distributed learning algorithms for training their models. Also, in most applications, the user data are generated and stored on the user-owned edge devices such as mobile phones and laptops. Today's security and utility requirements call for machine learning schemes that allow distributed training of the models on user devices.
In this regard, federated learning paves the way for the privacy-aware training of machine learning models by removing the necessity for storing the data centrally. The distributed training scheme provided by the federated learning algorithms enables the efficient training of the models by training the local models and sharing only the model parameters with a centralized server. Federated learning mechanism exploits the computational resources of the users to distribute the load from the server while enabling the users to train better models.

In some cases, clients in a federated learning problem may be subject to different objectives. By using federated multi-task learning methods, clients with different tasks and data distribution are able to cooperate in a federated learning environment.
It is also possible to bootstrap federated learning algorithms by employing transfer learning methods in federated learning problems. 

\section{Related Work}
The idea of federated learning was first presented in \cite{mcmahan2017communication} with the introduction of the FedAvg algorithm. FedAvg algorithm selects a random subset of the clients in each round and aggregates the model updates of the clients by averaging. Federated learning enables securely training machine learning models using distributed data while preserving the privacy of the users.
More information on federated learning can be found in the overview article \cite{li2020federated}, and the survey \cite{kairouz2021advances}.
In \cite{arivazhagan2019federated}, the authors proposed to add personalized layers to the models of each client and train those layers personally while training the base layers using the conventional federated learning scheme. In this way, they aimed to reduce the effects of statistical heterogeneity. That idea is similar to the technique proposed in this paper; however, we propose a more general framework and apply it to the multi-task learning problem.

The tasks of the clients are not necessarily the same in a federated learning environment. Clients with different tasks may cooperate for training a shared model in a federated environment. Even the source and target domains of the client datasets may differ in a federated learning setup.
A federated multi-task learning algorithm named MOCHA is introduced in \cite{smith2017federated} for convex models, such as \ac{SVM}.
The study in \cite{corinzia2019variational} proposes a federated multi-task learning algorithm for learning non-convex models.
A federated multi-task learning algorithm is introduced in \cite{mills2022multitask}. It enables personalized training of a federated deep neural network model by introducing batch normalization layers into the model. 

Another formulation for sharing model parameters in a federated learning algorithm is federated transfer learning.
A federated transfer learning framework is proposed for wearable healthcare in \cite{chen2020fedhealth}. The authors suggest a two-stage learning algorithm by performing conventional federated learning in the first stage and then fine-tuning personalized models in the next stage.
Another two-stage federated transfer learning for personalized model training is introduced by \cite{fallah2020personalized}. These authors try to find an initial model that can be fine-tuned by each client in the second stage.

Multi-task learning is a sub-area of transfer learning. However, it is generally referred to as a separate category due to its wide range of applications. Both fields can be grouped under the field of meta-learning. Meta-learning is a well-studied and active area of machine learning. More meta-learning methods could be adopted into the field of federated learning in the future. A comprehensive overview of meta-learning approaches can be found in \cite{hospedales2021meta}.

In a federated multi-task learning problem, the trained deep neural network model should be fine-tuned for the respective objective of each client while sharing some parameters for more generalizability. The first layers of a deep neural network learn more general features, while the deeper layers learn more specialized features. Following the properties of the networks mentioned above, we propose to train a deep neural network model with more generalized layers closer to the input and more personalized layers to the output. We achieve that by introducing layer types such as pre-trained, common, task-specific, and personal layers.

\subsection{Contributions}
The contributions of this paper can be summarized as follows:
\begin{itemize}
    \item We propose a novel generalized framework for multi-task federated learning applications. We propose a layered, hierarchical architecture for training federated learning models. We divide the layers of the deep learning model into four different groups: (i) The pre-trained layers, adopted from a pre-trained model trained on a similar task using the transfer learning methods, (ii) Common layers, trained using the conventional federated learning scheme, in which the parameters are shared by all the clients in the network, (iii) Task-specific layers, trained cooperatively by the clients having the same objective, (iv) Personal layers, trained separately by each client in the network. We demonstrated the proposed scheme using the deep neural network architecture as a machine learning model due to its layered structure. Furthermore, the proposed scheme is applicable to any machine learning model because it is a generalized framework.
    Although there are several studies in the literature that employs the idea of personal layers, clustered federated learning, or transfer learning, this paper is the first study to propose a generalized framework combining these ideas to the best of the authors' knowledge.
    
    \item We created a multi-label face dataset using the StyleGAN2\cite{karras2020analyzing} network. The image samples are labeled according to five different categories such as gender, age group, wearing glasses or not, pose angle, and the visibility of teeth. The proposed multi-task face dataset is used for demonstration of the advantages of the proposed federated multi-task learning framework. We performed extensive simulations for comparing the proposed framework to the cases such as separate or joint training of a centralized dataset, separate, or conventional federated learning training of the distributed dataset. We also further increase the performance of the proposed architecture by utilizing the transfer learning methods. As an example, we used the lower layers of the pre-trained VGG-Face\cite{ozbulak2016transferable} CNN model for our deep learning model.
    We also performed additional experiments on the human activity recognition dataset provided in \cite{anguita2013public}.
\end{itemize}

The rest of the paper is organized as follows. The proposed multi-task federated learning framework is introduced in \SEC{sec:mult-fed-learn}. Information about the generated face dataset is provided in \SEC{sec:dataset}. Extensive experiments for validating the performance of the proposed algorithm are presented in \SEC{sec:experiments}. Additional experiments corroborating the performance of the proposed algorithm on the human activity recognition dataset are provided in \SEC{sec:additional-experiments}. Finally, the conclusions are drawn in \SEC{sec:conclusion}.

\section{Proposed Multi-Task Federated Learning Framework}
\label{sec:mult-fed-learn}

A typical federated learning problem consists of $M$ distinct clients participating in a joint optimization problem without sharing their local data. The features and the labels in the dataset are denoted as $X$ and $Y$, respectively. The number of data samples for client $m$ is denoted as $N_M$, and the total number of data samples is indicated by $N$:
\begin{equation*}
    N = \sum_{m=1}^{M} N_M.
\end{equation*}
There are $T$ different labels/tasks in the dataset. In each round of the algorithm, $K$ clients are chosen randomly from each client group with the same task.  

\begin{figure*}[htbp]
    \vskip 0.2in
    \centering
    \includegraphics[width=0.8\linewidth]{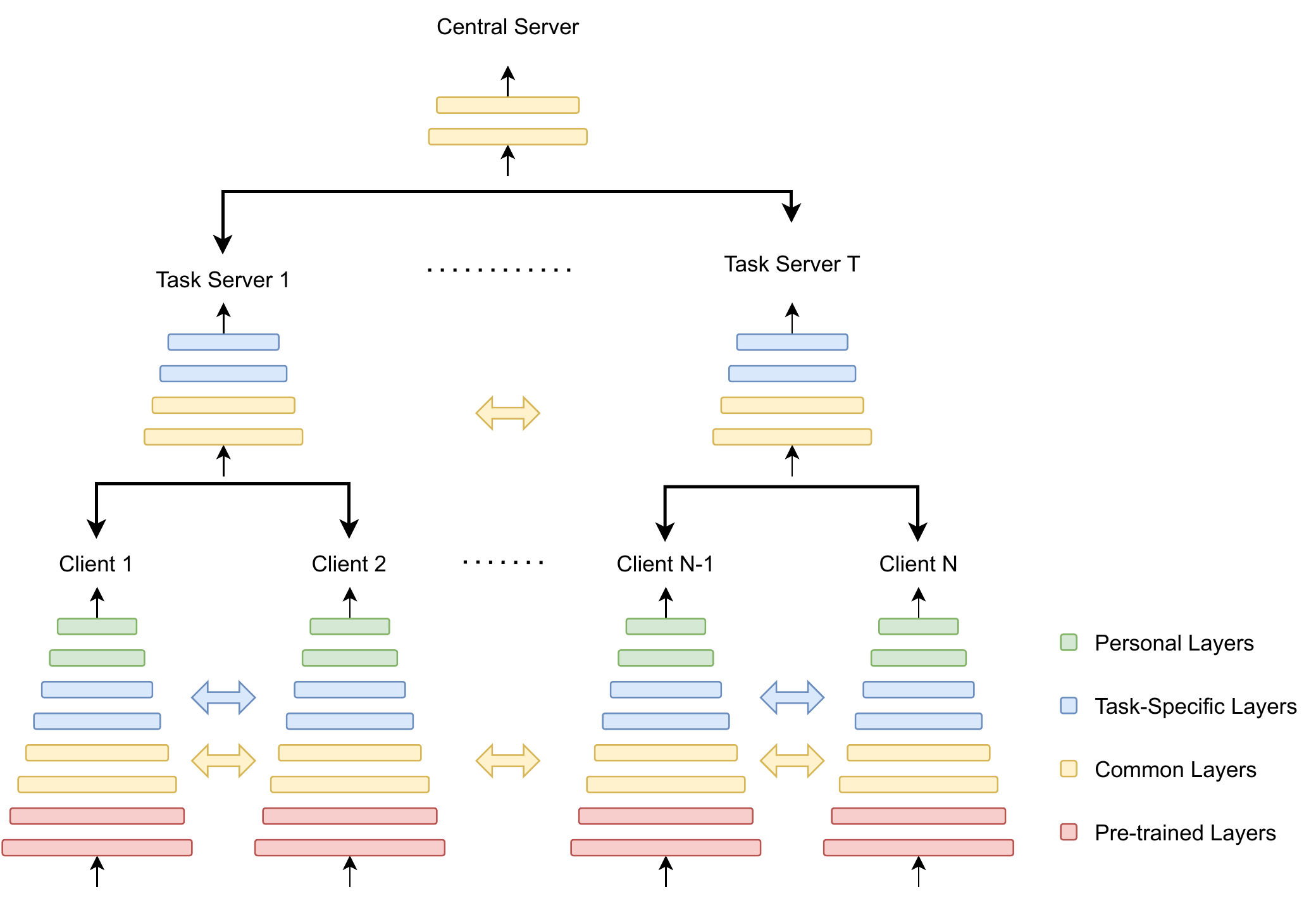}
    \caption{Architecture for the proposed multi-task federated learning model.}
    \label{fig:architecture}
    \vskip -0.2in
\end{figure*}

In this framework, we consider deep transfer learning, where we train a deep neural network model using the algorithms from both transfer and multi-task learning. The trained deep neural network model consists of four different layer types. There may be multiple layers for each layer type. In the lower layers of the model, there are pre-trained layers in order to utilize conventional transfer learning. Those layers may be adopted from a deep neural network model that was previously trained on a dataset that has a similar feature domain. The layers of the second type are the common layers that are connected to the output of the pre-trained layers. The weights of the common layers are shared by all the clients and trained using the conventional federated learning scheme. That is, the local updates of each client are aggregated in each round of the algorithm. The next layer type is the task-specific layers, which are trained separately for each client group. A client group consists of clients that share the same objective. The round updates for the task-specific layers are aggregated among the clients with the same task using federated learning algorithms. A separate server could be allocated for each task group in order to distribute the computational and communication load of the central server. The last layer type is the personal layer that is trained separately by each client. The architecture for the proposed algorithm is visualized in \FIG{fig:architecture}.

If we discard the group and unique layers, our model approaches conventional federated learning. Similarly, if we discard the common and group layers, it becomes separate training for each client. Hence, the proposed method is a generalized framework for combining the benefits of federated learning, deep transfer learning, and multi-task learning.

The weights of the model are denoted by $W$, which is partitioned into different layer types as follows
\begin{equation*}
    W = \left[ W_{\text{pre}}, W_{\text{com}}, W_{\text{task}}, W_{\text{pers}} \right]
\end{equation*}
where $W_{\text{pre}}$, $W_{\text{com}}$, $W_{\text{task}}$, and $W_{\text{pers}}$ represent the pre-trained, common, task specific, and personal layers, respectively.

Each layer group in the deep neural network model applies a function $h_{i}(x)$ to its input $x$. Function $h_{i}(x)$ is defined as
\begin{multline*}
    h_{i}(X) = \sigma_{L_{i}} \left( \sigma_{L_{i}-1} \left( \dots \sigma_{1} \left( X; W_{1} \right); W_{L_{i}-1} \right); W_{L_{i}} \right)\\
    \forall i \in \lbrace \text{pers},\text{task},\text{com},\text{pre}\rbrace.
\end{multline*}
Hence, the deep neural network can be represented in terms of the composite function $h(x)$:
\begin{equation*}
    h(X) = h_{\text{pers}} \left( h_{\text{task}} \left( h_{\text{com}} \left( h_{\text{pre}} \left( X \right) \right) \right) \right).
\end{equation*}
The objective is to minimize the loss function: 
\begin{equation*}
    W = \argmin_{W} \ell \left( h(X); Y \right).
\end{equation*}
where $\ell(\cdot)$ is an arbitrary loss function, such as categorical cross-entropy or mean square error.

By using the network-based deep transfer learning methods \cite{tan2018survey}, we may reuse a part of the layers of a pre-trained deep neural network to increase the performance of our model.
By employing this methodology, the weights of the pre-trained layers $W_{\text{pre}}$ are initialized by a pre-trained model and kept constant during our training. The other layers are trained using the \ac{SGD} algorithm, where the update is applied as in each round
\begin{equation*}
    W_{i} = W_{i} - \eta\nabla W_{i}, \quad \forall i 
\end{equation*}
where $\eta$ represents the learning rate.

The common layers are trained cooperatively by all the clients in the network. In each communication round, a subset of the clients are selected randomly and trained jointly. The gradients for the common layers are aggregated using the following formula
\begin{multline} \label{eq:agg-com}
    \nabla W_{\text{com},l} = \frac{1}{TKN} \sum_{t=1}^{T} \sum_{m=1}^{K} N_{m} \nabla W_{\text{com},l}^{(t,m)},\\
    \forall l=1,\dots,L.
\end{multline}
The task specific layers are trained separately for each task. The task specific layers of each task are jointly trained by the clients presenting that task. In each round, the selected clients in each task group updates the respective task specific layer. Hence, the gradient updates for the task specific layers are aggregated as
\begin{multline} \label{eq:agg-task}
    \nabla W_{\text{task},l}^{(t)} = \frac{1}{KN} \sum_{m=1}^{K} N_{m} \nabla W_{\text{task},l}^{(t,m)},\\
    \forall l=1,\dots,L, \forall t=1,\dots,T.
\end{multline}
Finally, each client trains its own personal layers. The personal layers are updated in each round by using the following recursion: 
\begin{equation*}
    W_{\text{pers},l}^{(m)}  = W_{\text{pers},l}^{(m)} - \eta\nabla W_{\text{pers},l}^{(m)}, \quad \forall m=1,\dots,K.
\end{equation*}

The aggregation method presented above illustrates the FedAvg algorithm. However, any federated aggregation method can be used to aggregate the client gradients. Additionally, this architecture provides the flexibility that different gradient aggregation methods could be employed for the common and task-specific layers when aggregating the weight updates.

\begin{algorithm}[htbp]
   \caption{Proposed algorithm}
   \label{alg:proposed}
\begin{algorithmic}
    \STATE {\bfseries Input:} dataset $X,Y$, size $m$
    \FOR{rounds $r=1$ {\bfseries to} $R$}
        \STATE broadcast the common weights $W_{\text{com}}$
        \FOR{tasks $t=1$ {\bfseries to} $T$}
            \STATE select $K$ clients randomly
            \STATE broadcast the task-specific weights $W_{\text{task}}^{(t)}$
            
            \FOR{clients $m=1$ {\bfseries to} $K$}
                \STATE compute the local gradients $\nabla W^{(m)}$
                \STATE apply the personal layer updates $\nabla W_{\text{pers}}^{(m)} $
            \ENDFOR
            \STATE aggregate and apply task-specific gradients $\nabla W_{\text{task}}^{(t)}$ using \eqref{eq:agg-com}
        \ENDFOR
        \STATE aggregate and apply common gradients $\nabla W_{\text{com}}$ using \eqref{eq:agg-task}
    \ENDFOR
\end{algorithmic}
\end{algorithm}

\section{Dataset}
\label{sec:dataset}
\begin{figure}[htbp]
    \vskip 0.2in
    \begin{center}
        \centerline{\includegraphics[width=\columnwidth]{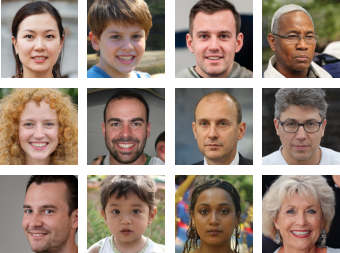}}
        \caption{Sample images from the dataset.}
        \label{fig:sample-images}
    \end{center}
    \vskip -0.2in
\end{figure}

We generated a face dataset\footnote{The dataset will be publicly available in the publication stage.} using StyleGAN2 network proposed in \cite{karras2020analyzing}. Image samples from the dataset are shown in \FIG{fig:sample-images}.
We labeled each image for five features according to gender, age group, whether wearing glasses, pose angle, and whether the teeth are visible.
The labels are summarized in \TAB{tab:dataset-features}. There are 13792 samples in the dataset.
The distribution of the label values for each feature is provided in \TAB{tab:dataset-distribution}.

\begin{table}[htbp]
	\caption{Features in the dataset.}
	\label{tab:dataset-features}
	\vskip 0.15in
	\begin{center}
	\begin{small}
	\begin{sc}
	\begin{tabular}{lcc}
		\hline
		Feature & 0 & 1 \\
		\hline
		 Gender     & Male        & Female       \\
		 Age Group  & Child       & Adult        \\
		 Glasses    & Not wearing & Wearing      \\
		 Pose Angle & Sideways    & Front-facing \\
		 Teeth      & Not showing & Showing      \\
		\hline
	\end{tabular}
	\end{sc}
	\end{small}
	\end{center}
	\vskip -0.1in
\end{table}

\begin{table}[htbp]
	\caption{Distribution of the labels in the dataset.}
	\label{tab:dataset-distribution}
	\vskip 0.15in
	\begin{center}
	\begin{small}
	\begin{sc}
	\begin{tabular}{lcc}
		\hline
		Feature & 0 & 1 \\
		\hline
		 Gender     & 6140 (44.52\%)  & 7652 (55.48\%)  \\
		 Age Group  & 1587 (11.51\%)  & 12205 (88.49\%) \\
		 Glasses    & 11477 (83.21\%) & 2315 (16.79\%)  \\
		 Pose Angle & 7115 (51.59\%)  & 6677 (48.41\%)  \\
		 Teeth      & 5224 (37.88\%)  & 8568 (62.12\%)  \\
		\hline
	\end{tabular}
	\end{sc}
	\end{small}
	\end{center}
	\vskip -0.1in
\end{table}

\section{Experiments}
\label{sec:experiments}

We performed the simulations using the face dataset mentioned in \SEC{sec:dataset}.
There are 100 clients in the system and five client groups corresponding to five different labels in the dataset. Hence, we have 20 clients per group.
We performed the simulations using Python programming language and Tensorflow framework \cite{tensorflow2015-whitepaper}.
We trained the models in four different setups in order to analyze the performance gain introduced by the proposed federated learning architecture.

The \ac{CNN} models trained on face images are shown to be transferable for soft biometric traits in \cite{ozbulak2016transferable}. The authors showed that VGG-Face network fine-tuned for age and gender classification performs better than the \ac{CNN} models trained from scratch. 
Following that idea, we used the VGG-Face network \cite{parkhi2015deep} for transfer learning for the pre-trained layers. The VGG-Face network has the VGG-16 architecture proposed in \cite{simonyan2014very}. The VGG-Face model was trained on 2.6M face images. The input shape is (224,224,3) for the VGG-Face architecture. We discard the last fully connected layer and connect the output of the fully connected layer 7 (FC7), which has an output dimension of 4096, to the input of the fine-tuning layers.
We use fully connected layers for fine-tuning the models. We use \ac{ReLU} activation function for the hidden layers and sigmoid activation function for the output layer.

In order to show the performance gain introduced by the proposed federated learning architecture, we compare the accuracy of five different cases. In the first two cases, the data is centralized, and we train (i) a single model with five outputs corresponding to each label and (ii) separate models for each of the five labels. In the next three cases, we distribute the data to separate clients and train (i) a separate model for each client, (ii) a separate model for each task in a federated learning scheme, (iii) using the proposed federated multi-task learning approach.

We use four additional feed-forward layers for fine-tuning the deep neural network model, including the output layer. The number of nodes are 4096, 2048, and 1024 for the hidden layers, respectively. The model is optimized using the \ac{SGD} algorithm, where the learning rate is selected via a multiplicative grid for each case. For the federated learning cases, the number of local epochs and the number of active clients per round are also optimized using a parameter grid. The batch size used in training is 16.

\subsection{Centralized Separate Training}
In this case, we train the neural network model using the whole dataset in a centralized machine. A separate model is trained for each task. The dataset size is the same for each model.

\subsection{Centralized Joint Training}
In this case, we train the neural network model using the whole dataset in a centralized machine. We train a single model with each task as a node in the output layer. We use the sigmoid activation function in the output layer in order to train for each label independently.

\subsection{Distributed Separate Training}
In this case, the dataset is distributed to 100 clients. The dataset is distributed uniformly to the clients. The clients are assigned uniformly into five different clusters corresponding to each task in the dataset. Each client has the label for only the task it is assigned to.
We consider this case to show the incentive for each client to participate in a federated learning scheme.

\subsection{Distributed Separate Federated Training}
In this case, the dataset is distributed similarly to the previous case. We performed separate federated training for each task. All four fine-tuning layers are trained using federated learning inside each task group. Hence, there are five distinct models in this scenario.

\subsection{Distributed Federated Multi-Task Training}
In this case, we implement the proposed method. The dataset is distributed to the clients similar to the previous case and trained using the proposed federated multi-task training architecture.
The first two fine-tuned layers are assigned to the commonly trained layers, and the other two (including the output layer) are assigned to the task-specific layers. No personally trained layer is used in the simulations.

\begin{figure}[htbp]
    \vskip 0.2in
    \begin{center}
        \centerline{\includegraphics[width=\columnwidth]{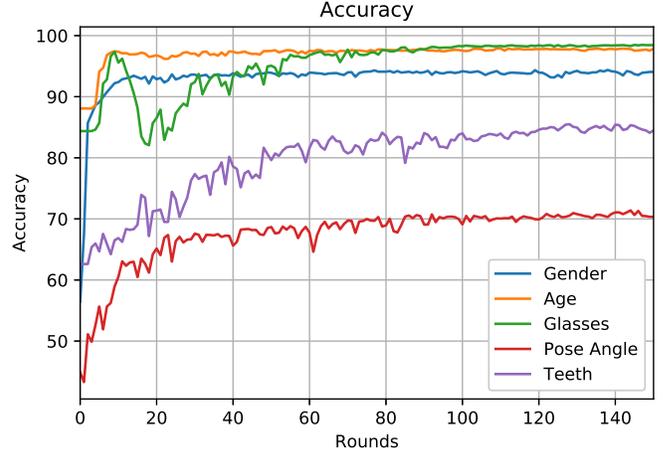}}
        \caption{Change of the test accuracy (\%) through the rounds for the federated multi-task training.}
        \label{fig:fed-acc}
    \end{center}
    \vskip -0.2in
\end{figure}

The test accuracy values for each round are plotted in \FIG{fig:fed-acc}.
The comparison of the test accuracy for each case is provided in \TAB{tab:accuracy-comparison}.

The accuracy values of the centralized training cases are expected to show an upper bound on the accuracy values of the federated learning cases since the whole dataset is available for the training.
As expected, the performance of the centralized dataset cases is the highest for each task. There is no significant difference in the accuracy values for the centralized separate and joint training cases.
The performance in the distributed data and separate training case is the worst because the models are trained with a small amount of local data by each client.
The accuracy values get significantly higher for each task in both federated learning scenarios. This performance gap may provide an incentive for a client to participate in a federated learning environment.
For our dataset, there is not a significant performance difference between the separate and multi-task federated learning cases. However, the federated multi-task learning algorithm has the advantage that only a single model needs to be trained.

\begin{table*}[tb]
	\caption{Comparison of the accuracy (\%) values for each case for each task.}
	\label{tab:accuracy-comparison}
	\vskip 0.15in
	\begin{center}
	\begin{small}
	\begin{sc}
	\begin{tabular}{lccccc}
		\hline
		Case & Gender & Age Group & Glasses & Pose Angle & Teeth \\
		\hline
		 Centralized Separate & 95.32 & 97.93 & 99.38 & 77.12 & 88.11 \\
		 Centralized Joint & 95.29 & 97.98 & 99.35 & 77.14 & 87.75 \\
		 Distributed Separate & 91.64 & 96.22 & 95.49 & 60.53 & 73.30 \\
		 Distributed Separate FL & 94.52 & 97.90 & 99.02 & 72.48 & 86.80 \\
		 Distributed Multi-Task FL & 94.56 & 97.90 & 98.66 & 73.02 & 86.08 \\
		\hline
	\end{tabular}
	\end{sc}
	\end{small}
	\end{center}
	\vskip -0.1in
\end{table*}

\section{Additional Experiments}
\label{sec:additional-experiments}

In this section, we apply the proposed multi-task federated learning framework to the human activity recognition dataset provided in \cite{anguita2013public}.
The dataset was constructed by accelerometer and gyroscope measurements obtained through the mobile phones of the subjects. The samples are labeled by one of the six activities performed by the subjects: (i) walking, (ii) walking upstairs, (iii) walking downstairs, (iv) sitting, (v) standing, (vi) laying.
We modeled each subject as a different task. Hence, we have 30 tasks and one client per each task.
We used the feature vectors with the length of 561 provided in \cite{anguita2013public}.
In this experiment, we do not employ pre-trained layers since there is no available pre-trained model for the tasks. Since we model each subject as a task, the task layers and the personal layers are merged.

\begin{table}[tb]
	\caption{Comparison of the accuracy (\%) values for each case for the human activity recognition.}
	\label{tab:har-accuracy}
	\vskip 0.15in
	\begin{center}
	\begin{small}
	\begin{sc}
	\begin{tabular}{lccccc}
		\hline
		Case & Accuracy \\
		\hline
		 Separate & 98.89 \\
		 Centralized Joint & 98.98 \\
		 Multi-Task FL & 99.42 \\
		\hline
	\end{tabular}
	\end{sc}
	\end{small}
	\end{center}
	\vskip -0.1in
\end{table}

\begin{figure}[htbp]
    \vskip 0.2in
    \begin{center}
        \centerline{\includegraphics[width=\columnwidth]{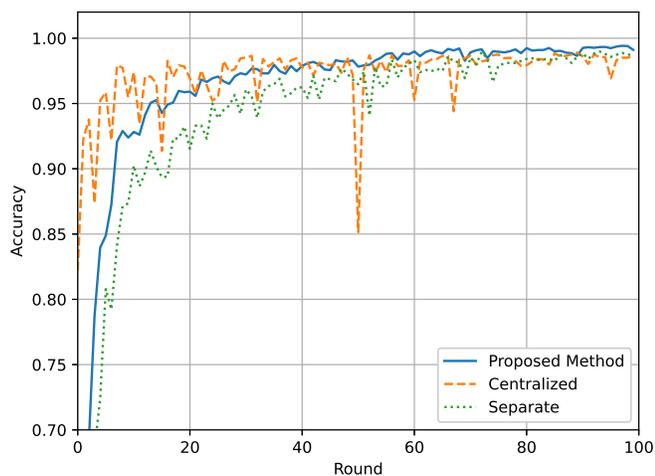}}
        \caption{Change of the test accuracy (\%) through the rounds for the training of the human activity recognition dataset.}
        \label{fig:uci-har-acc}
    \end{center}
    \vskip -0.2in
\end{figure}

In this experiment, the dataset of each subject has the same objective, which is the label of their physical activity; however, the distribution of the features is not the same. Hence, the classification tasks of each subject are modeled separately.
We used a multi-layer perceptron as a deep learning architecture. We fixed the number of hidden layers and the number of nodes on each layer for all scenarios. The number of hidden dense layers is three, with the number of nodes 128, 64, and 32, respectively. We used \ac{SGD} as an optimizer and optimized the learning rates and the batch size using a parameter grid.
We trained the deep learning model in three different scenarios for comparison. In the first scenario, we trained a single model using the whole dataset, i.e., the datasets of the subjects are combined. In the second scenario, we trained a separate model for the dataset of each subject. Hence, we obtained 30 different models. Finally, we trained the deep learning classifier using the proposed federated multi-task learning framework.

The accuracy values for each scenario are shown in \TAB{tab:har-accuracy}.
Due to the difference in the underlying data distributions, the trained centralized joint training performs the worst among the scenarios we considered. On the other hand, separate training of the models for each subject gives us better results in terms of classification accuracy. The proposed federated multi-task learning framework provides the best results since it enables personalized training for the tasks. The change of the test accuracy for each scenario is provided in \FIG{fig:uci-har-acc}. Although the centralized training provides a faster convergence speed at the first rounds, it converges to a smaller accuracy value. On the other hand, the separate training converges slowly but to a higher accuracy value than the centralized training. The proposed method provides convergence to a higher accuracy value.

\section{Conclusion}
\label{sec:conclusion}

In this paper, we have shown that clients with different tasks but similar dataset domains could cooperate in a federated learning scheme in order to improve their performance. We compared different methods for training a multi-task objective using both centralized and distributed datasets. We concluded that combining both transfer learning and multi-task learning in a federated learning environment could improve the performance of a deep neural network classifier.

% TODO: Crop the long reference

\bibliographystyle{IEEEtran}
\bibliography{main}

\end{document}